\documentclass[runningheads]{llncs}
\usepackage{graphicx}
\usepackage{comment}

\usepackage{amsmath,amssymb} 
\usepackage{color}

\usepackage[width=122mm,left=12mm,paperwidth=146mm,height=193mm,top=12mm,paperheight=217mm]{geometry}
\usepackage{tabularx}
\usepackage{multirow}
\usepackage{amsfonts}
\usepackage{subfigure}
\begin{document}
\pagestyle{headings}
\mainmatter
\def\ECCVSubNumber{1295}  

\title{Unsupervised Deep Metric Learning with Transformed Attention Consistency  and Contrastive Clustering Loss} 

\titlerunning{Unsupervised Deep Metric Learning with TAC and CCL}
%
\author{Yang Li\inst{1}\and
Shichao Kan\inst{2} \and 
Zhihai He\inst{1}}
\authorrunning{Y. Li, S. Kan, and Z. He}
%
\institute{University of Missouri, Columbia, MO, USA\\
\email{yltb5@mail.missouri.edu,  hezhi@missouri.edu}\\
\and
Beijing Jiaotong University, Beijing, China\\
\email{16112062@bjtu.edu.cn}\\
}
\maketitle

\begin{abstract}

Existing approaches for unsupervised metric learning focus on
exploring self-supervision information within the input image itself. We observe that, when analyzing images, human eyes often compare images against each other instead of examining images individually. In addition, they often pay attention to certain keypoints, image regions, or objects which are discriminative between image classes but highly consistent within classes. Even if the image is being transformed, the attention pattern will be consistent. 
Motivated by this observation, we develop a new approach to unsupervised deep metric learning where the network is learned based on self-supervision information across images instead of
within one single image. To characterize the consistent pattern of human attention during image comparisons, we introduce the idea of transformed attention consistency. It assumes that visually similar images, even undergoing different image transforms, should share the same consistent visual attention map. This consistency leads to a pairwise self-supervision loss, allowing us to learn a Siamese deep neural network to encode and compare images against their transformed or matched pairs. To further enhance the inter-class discriminative power of the feature generated by this network, we adapt the concept of triplet loss from supervised metric learning to our unsupervised case and introduce the contrastive clustering loss. Our extensive experimental results on  benchmark datasets demonstrate that our proposed method outperforms current state-of-the-art methods for unsupervised metric learning by a large margin.


\end{abstract}

\section{Introduction}

Deep metric learning  aims to learn discriminative features that can aggregate visually similar images into compact clusters in the high-dimensional feature space while separating images of different classes from each other.
It  has many important applications, including image retrieval \cite{wohlhart2015learning,he2018triplet,grabner20183d}, face recognition \cite{wen2016discriminative}, visual tracking \cite{tao2016siamese} and person re-identification \cite{yu2018hard,hermans2017defense}.
In supervised deep metric learning, we assume that the labels for training data are available. 
In this paper, we consider unsupervised deep metric learning where the image labels are not available.
Learning directly and automatically from images in an unsupervised manner without human supervision represents 
a very important yet challenging task in computer vision and machine learning.

Clustering is one of the earliest methods developed for unsupervised learning. 
Recently, motivated by the remarkable success of deep learning, researchers have started to develop unsupervised learning methods using deep neural networks \cite{caron2018deep}. Auto-encoder trains an encoder deep neural network to output feature representations with sufficient information to reconstruct input images by a paired decoder \cite{zhang2019aet}. 
As we know, during deep neural network training, the network model is  updated and learned in an iterative and progressive manner so that the network output can match the target. In other works, deep neural networks need human supervision to provide ground-truth labels. However, in unsupervised learning, there are no labels available. To address this issue, researchers have exploited the unique characteristics of images and videos to create various self-supervised labels, objective functions, or loss functions, which essentially convert the unsupervised learning into a supervised one so that the deep neural networks can be successfully trained. 
For example, in DeepCluster \cite{caron2018deep},  clustering is used to generate pseudo labels for images. 
Various supervised learning methods have been developed to train networks to predict the relative position of two randomly sampled patches \cite{doersch2015unsupervised}, 
solve Jigsaw image puzzles \cite{noroozi2016unsupervised}, predict pixel values of missing image patches \cite{devries2017improved}, classify image rotations of four discrete angles \cite{gidaris2018unsupervised}, reconstruct image transforms 
\cite{zhang2019aet}, etc. 
Once successfully trained by these pretext tasks, the baseline network should be able to generate discriminative features for subsequent tasks, such as image retrieval, classification, matching, etc \cite{he2019momentum}.

\begin{figure}[h]
	\begin{center}
		\includegraphics[width=0.85\linewidth]{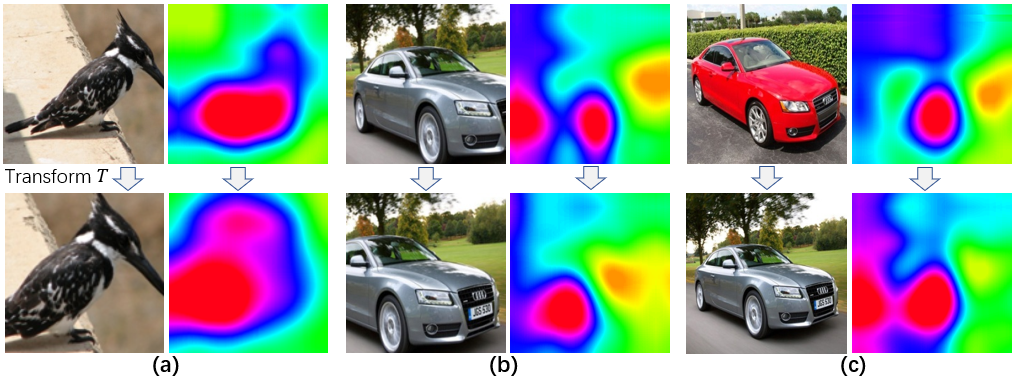}
	\end{center}
	\vspace{-6mm}
	\caption{Consistency of visual attention across images under transforms.}
	\label{fig:transformed-attention}
	\vspace{-4mm}
\end{figure}

In this work, we propose to explore a new approach to unsupervised deep metric learning. We observe that existing methods for unsupervised metric learning focus on learning a network to analyze the input image itself.  As we know, when examining and classifying images, human eyes compare images back and forth in order to identify discriminative features \cite{gazzaniga2009cognitive}.  In other words, comparison plays an important role in human visual learning.
When comparing images, they often pay attention to certain keypoints, image regions, or objects which are discriminative between image classes but highly consistent across image within classes. 
Even when the image is being transformed, the attention areas will be consistent. To further illustrate this, we provide three examples in Fig. \ref{fig:transformed-attention}. In (a), human eyes can easily tell the top image $A$ of the first column and the bottom image $B$ are  the same bird since they have the same visual characteristics. The attention will be on the feather texture and head shape. In the pixel domain, $A$ and $B$ are up to a spatial transform, specifically, cropping plus resizing.
When the human eyes moves from image $A$ to its transformed version $B$, the attention will be also transformed so that it can be still focused on the head and feather. If we represent this attention using the attention map in deep neural networks, the attention map ${\cal M}(A)$ for image $A$ and the attention map ${\cal M}(B)$ for image $B$ should also follow the same transform, as shown in the second column of Fig. \ref{fig:transformed-attention}(a). We can also see this consistency of attention across image under different transforms in other examples in Figs. \ref{fig:transformed-attention} (b) and (c). 

This lead to our idea of transformed attention consistency. Based on this idea, we develop a new approach to unsupervised deep metric learning based on image comparison. Specifically, using this consistency, we can define   a pairwise self-supervision loss, allowing us to learn a Siamese deep neural network to encode and compare images against their transformed or matched pairs. To further enhance the inter-class discriminative power of the feature generated by this network, we adapt the concept of triplet loss from supervised metric learning to our unsupervised case and introduce the contrastive clustering loss. Our extensive experimental results on  benchmark datasets demonstrate that our proposed method outperforms current state-of-the-art methods by a large margin. 

\section{Related Work and Major Contributions}
This work is related to deep metric learning, self-supervised representation learning, unsupervised metric learning, and attention mechanisms. 

\textbf{(1) Deep metric learning.}
The main objective of deep metric learning is to learn a non-linear transformation of an input image by deep neural networks. In a common practice \cite{wang2019multi,ye2019unsupervised}, the backbone in deep metric learning can be pre-trained on 1000 classes ImageNet \cite{russakovsky2015imagenet} classification, and is then jointly trained on the metric learning task with an additional linear embedding layer. Many recent deep metric learning methods are built on pair-based \cite{chopra2005learning,hadsell2006dimensionality,oh2016deep} and triplet relationships \cite{kumar2017smart,schroff2015facenet,wu2017sampling}.
Triplet loss \cite{schroff2015facenet} defines a positive pair and a negative pair based on the same anchor point. It encourages the embedding distance of positive pair to be smaller than the distance of negative pair by a given margin. 
Multi-similarity loss \cite{wang2019multi} considers multiple similarities and provides a more powerful approach for mining and weighting informative pairs by considering multiple similarities.
The ability of mining informative pairs in existing methods is limited by the size of mini-batch.
Cross-batch  memory (XBM) \cite{wang2019cross} provides a memory bank for the feature embeddings of past iterations. In this way, the informative pairs can be identified across the dataset instead of a mini-batch. 

\textbf{(2) Self-supervised representation learning.}
Self-supervised representation learning directly derives information from unlabeled data itself by formulating predictive tasks to learn informative feature representations.  
DeepCluster \cite{caron2018deep} uses $k$-means clustering to assign pseudo-labels to the features generated by the deep neural network and introduces a discriminative loss to train the network. 
Gidaris \textit{et al.} \cite{gidaris2018unsupervised} explore the geometric transformation and propose to predict the angle ($0^\circ$, $90^\circ$, $180^\circ$, and $270^\circ$) of image rotation as a four-way classification.
Zhang \textit{et al.} \cite{zhang2019aet} propose to predict the randomly sampled transformation from the encoded features by Auto-encoding transformation (AET). The encoder is forced to extract the features with visual structure information, which are informative enough for the decoder to decode the transformation. 
Self-supervision has been widely used to initialize and pre-train backbone on unlabeled data, and is then fine-tuned on a labeled training data for evaluating different tasks.

\textbf{(3) Unsupervised metric learning.}
Unsupervised metric learning is a relatively new research topic. It is a more challenging task since the training classes have no  labels and it does not overlap with the testing classes.
Iscen \textit{et al.} \cite{iscen2018mining} propose an unsupervised method to mine hard positive and negative samples based on manifold-aware sampling. The feature embedding can be trained with standard contrastive and triplet loss.
Ye \textit{et al.} \cite{ye2019unsupervised} propose to utilize the instance-wise relationship instead of class information in the learning process. It optimizes the instance feature embedding directly based on the positive augmentation invariant and negative separated properties. 

\textbf{(4) Attention mechanism.}
The goal of the attention mechanism is to capture the most informative feature in the image. It explores important parts of features and suppress unnecessary parts  \cite{ba2014multiple,mnih2014recurrent,jaderberg2015spatial}.
Convolutional block attention module (CBAM) \cite{woo2018cbam} is an effective attention method with channel and spatial attention module which can be integrated into existing convolutional neural network architectures. 
Fu \textit{et al.} \cite{fu2017look} propose to produce the attention proposals  and train the attention module and embedding module in an iterative two-stage manner. Chen \textit{et al.} \cite{chen2019hybrid} propose the hybrid-attention system by random walk graph propagation for object attention and the adversary constraint for channel attention. 

Compared to existing methods, the \textbf{\textit{unique contributions}} of this paper can be summarized as follows. (1) Unlike existing methods which focus on information analysis of the input image only, we explore a new approach for unsupervised deep metric learning based on image comparison and cross-image consistency.
(2) Motivated by the human visual experience, we introduce the new approach of transformed attention consistency to effectively learn a deep neural network which can focus on discriminative features. 
(3) We extend the existing triplet loss developed for supervised metric learning to unsupervised learning using $k$-mean clustering to assign pseudo labels and memory bank to allow its access to all training samples, instead of samples in the current mini-batch. 
(4) Our experimental results demonstrate that our proposed approach has improved the state-of-the-art performance by a large margin.


\begin{figure}[h]
	\begin{center}
		\includegraphics[width=0.8\linewidth]{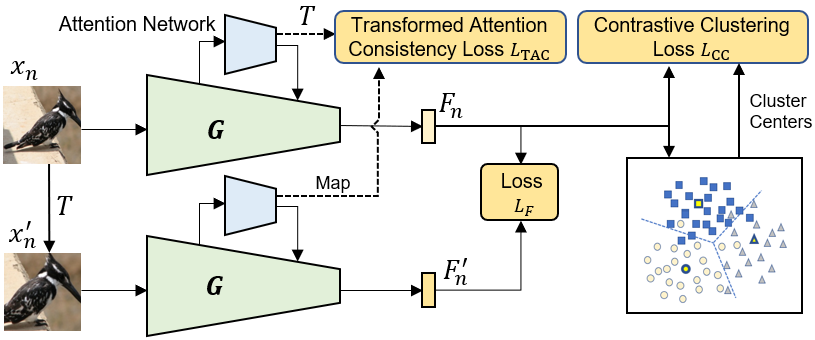}
	\end{center}
	\vspace{-6mm}
	\caption{Overview of the proposed approach for unsupervised deep metric learning with transformed attention consistency and contrastive clustering loss.}
	\label{fig:framework}
	\vspace{-4mm}
\end{figure}

\section{Method}

\subsection{Overview}
Suppose that we have a set of unlabeled images ${\cal X}=\{x_1, x_2, ..., x_N\}$. Our goal is to learn a deep neural network to extract their features  ${\mathbf{G}}(x_n) \in \mathbb{R}^d$, where $d$ is the feature dimension. 
Fig. 2 shows the overall design of our proposed method for unsupervised deep metric learning based on transformed attention consistency and contrastive clustering loss (TAC-CCL). Given an input image $x_n$, we apply a transform $T$, which is randomly sampled from a set of image transforms ${\cal T}$, to $x_n$, to obtain its transformed version
$x'_n = T(x_n)$. In our experiments, we mainly consider spatial transforms, including cropping (sub-image), rotation, zooming, and perspective transform. Each transform is controlled by a set of transform parameters. For example, the cropping is controlled by its bounding box. The perspective transform is controlled by its 6 parameters.
Image pairs $(x_n, x'_n)$ are inputs to the Siamese deep neural network. These two identical networks will be trained to extract features $F_n$ and $F'_n$ for these two images. As illustrated in Fig. 2, each network is equipped with an attention network to learn the attention map which will modulate the output feature map. 
The attention map can enforce the network to focus on discriminative local features for the specific learning tasks or datasets. 
Let $M_n$ and $M'_n$ be the attention maps for images $x_n$ and $x'_n$, respectively. According to the transformed attention consistency, we shall have 
\begin{equation}
    M'_n = T(M_n).
    \label{eq-mapping}
\end{equation}
Based on this  constraint, we introduce the transformed attention consistency loss $L_{TAC}$ to train the feature embedding network $\mathbf{G}$, which will be further explained in Section \ref{sec-loss}. 
Besides this attention consistency, we also require that the output features $F_n$ and $F_{n}^{'}$ should be similar to each other since the corresponding input images $x_n$ and $x_{n}^{'}$ are visually the same. 
To enforce this constraint, we introduce the feature similarity loss $L_F = ||{F_n - F_{n}^{'}}||_2$ which is the $L_2$-normal between these two features. 
To ensure that image features from the same class aggregate into compact clusters while image features from different classes are pushed away from each other in the high-dimensional feature space, we introduce the contrastive clustering loss $L_{CC}$, which will be further explained in the Section \ref{sec-loss}.

\subsection{Baseline System}
In this work, we first design a baseline system. 
Recently, a method called multi-similarity (MS) loss \cite{wang2019multi} has been developed for supervised deep metric learning. In this work, we adapt this method from supervised metric learning to unsupervised metric learning 
 using $k$-means clustering to assign pseudo labels. Also, the original MS method computes the similarity scores between image samples in the current mini-batch. 
In this work, we extend this similarity analysis to the whole training set using the approach of memory bank \cite{wang2019cross}.  
The features of all training samples generated by the network  are stored in the memory bank by the enqueue method. When the memory bank is full, the features and corresponding labels of the oldest mini-batch are removed by the dequeue method. Using this approach, the current mini-batch has access to the whole training set. We can then compute the similarity scores between all samples in the mini-batch and all samples in the training set. Our experiments demonstrate that this enhanced similarity matrix results in significantly improved  performance in unsupervised metric learning. 
In this work, we use this network as the baseline system,
denoted by TAC-CLL (baseline).

\subsection{Loss Functions}
\label{sec-loss}
To further improve the performance of the baseline system, we introduce the ideas of transformed attention consistency and contrastive clustering loss, which are explained in the following. 

The transformed attention consistency aims to enforce the feature embedding network $\mathbf{G}$ to focus visually important features instead of other background noise. Let $M_{n}(u, v)$ and $M_{n}^{'}(u, v)$ be the attention maps for input image pair $x_{n}$ and $x_{n}^{'}$, where 
$(u, v)$ represents a point location in the attention map.
Under the transform $T$, this point is mapped to a new location denoted by $(T_u(u, v), T_v(u, v))$
According to (\ref{eq-mapping}), if we transform the attention map $M_n(u, v)$ for the original image $x_n$ by $T$, it should match the 
attention map $M'_n(u, v)$ for the transformed image $x'_n = T(x_n)$. 
Based on this, the proposed transformed attention consistency loss $L_{TAC}$ is defined as follows 
\begin{equation}
    L_{TAC} = \sum_{(u, v)} |M_n(u, v) - M'_n(T_u(u, v), T_v(u, v))|^2,
\end{equation}
where $u'=T_u(u, v)$ and $v'=T_v(u, v)$ are the mapped location of $(u, v)$ in image $x'_n$. 

\begin{figure}[h]
    \begin{center}
  \hspace{-0mm}\subfigure[Baseline without TAC-CLL]{
			\label{fig:subfig_as1_a1}
        \includegraphics[width=0.38\linewidth]{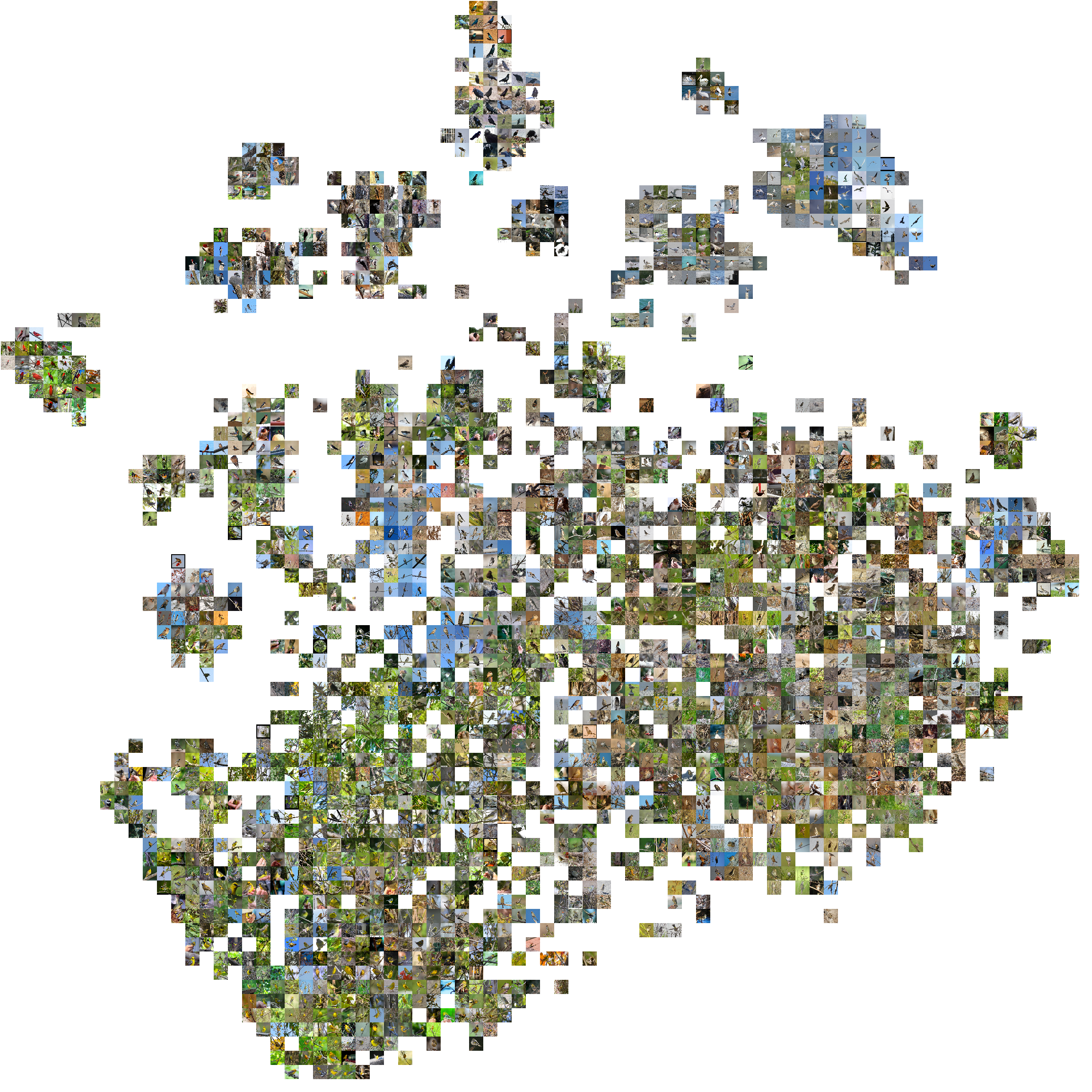}
		}
	\hspace{-0mm}\subfigure[Baseline with TAC-CCL]{
			\label{fig:subfig_as1_b1}
        \includegraphics[width=0.38\linewidth]{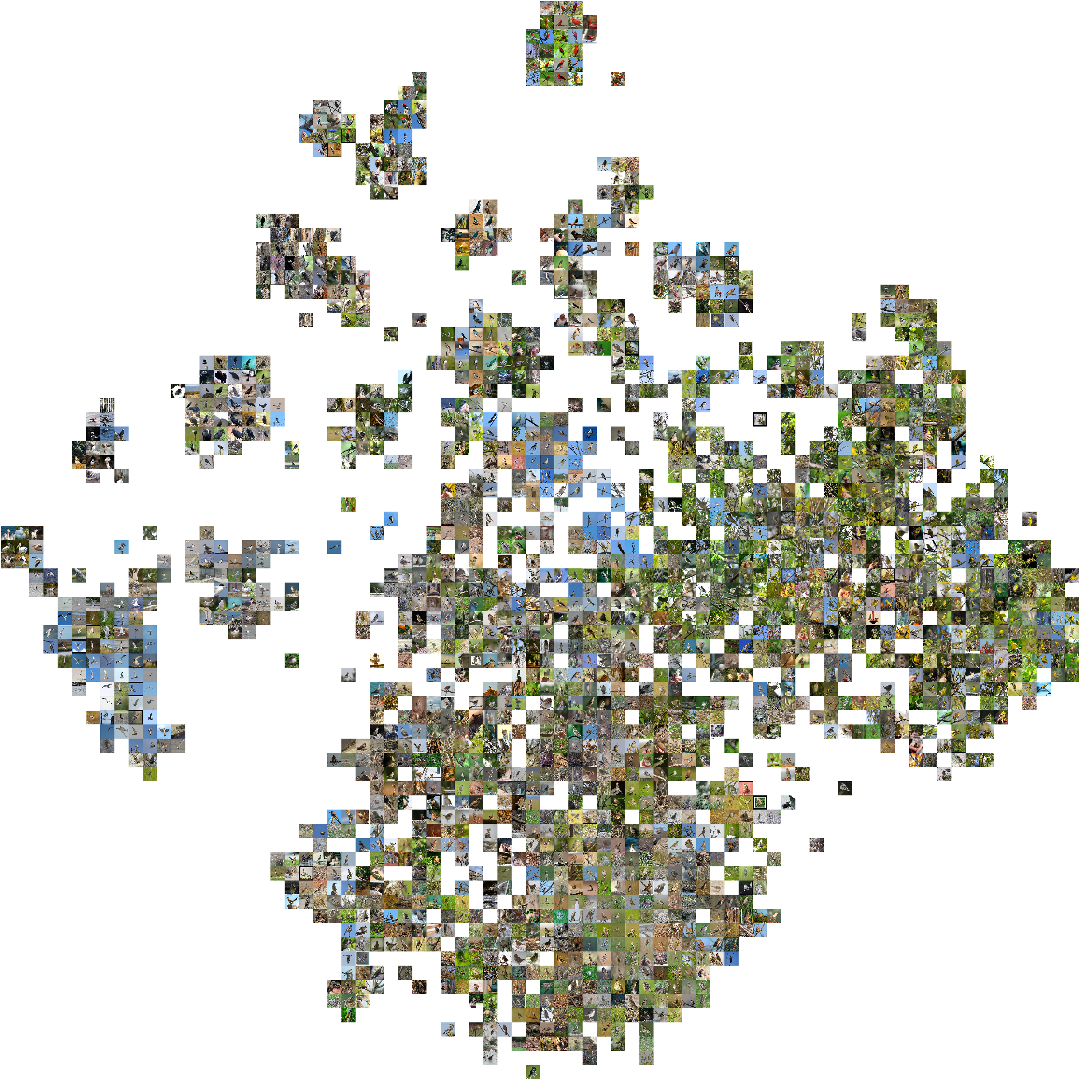}
		}
    \end{center}
    \vspace{-8mm}
	\caption{The Barnes-Hut t-SNE visualizations of the CUB test dataset with and without the TAC-CCL method.}
	\label{fig:tsne}
	\vspace{-1mm}
\end{figure}

The constrastive clustering loss extends the triplet loss \cite{schroff2015facenet} developed in supervised deep metric learning, where an anchor sample $x$  is associated with a positive sample $x_+$ and a negative sample $x_-$. The triplet loss aims to maximize the ratio  $S(x, x_+)/S(x, x_-)$, where $S(\cdot, \cdot)$ represents the cosine similarity between two features. It should be noted that this triplet loss requires the knowledge of image labels, which however are not available in our unsupervised case. To extend this triplet loss to unsupervised metric learning, we propose to cluster the image features into $K$ clusters. In the high-dimensional feature space, we wish these clusters are compact and are well separated from each other by large margins.
Let $\{{C}_k\}$, 1 $\le k \le$ $K$, be the cluster centers. Let $C_{+}(F_n)$ be the nearest center which has the minimum distance to the input image feature $F_n$ and the corresponding distance is denoted by $d_{+}(F_n)=||F_n - C_{+}(F_n)||_2$. Let 
$C_-(F_n)$ be the cluster center which has the second minimum distance to $F_n$ and the corresponding distance is denoted by $d_-(F_n)=||F_n - C_-(F_n)||_2$.
If the contrastive ratio of $d_+(F_n) / d_+(F_n)$ is small, then this feature has more discriminative power. We define the following contrastive clustering loss
\begin{equation}
    L_{CC} = \mathcal{E}_{F_n}\left\{
    \frac{||F_n - C_+(F_n)||_2}{||F_n - C_-(F_n)||_2}
    \right\},
\end{equation}
which is the average contrastive ratio of all input image features.
During the training process, the network $\mathbf{G}$, as well as the feature for each input, is progressively updated. For example, the clustering is performed and the cluster centers are updated for every 20 epochs. 

Fig. \ref{fig:tsne} visualizes the images of the CUB dataset in the feature space with features extracted by the baseline system with and without the TAC-CCL approach. We can see that using the TAC-CCL, the obtained feature clusters are more compact within each class and better separated from each other between classes.

\subsection{Transformed Attention Consistency with Cross-Images Supervision}

Note that, in our proposed approach, we transform or augment the input image $x_n$ to create its pair $x_{n}^{'}$. These two are from the same image source. 
We also notice that most of existing self-supervision methods, such as 
predicting locations of image patches and classifying the rotation of an image \cite{gidaris2018unsupervised}, and reconstructing the transform of the image \cite{zhang2019aet}, all focus on self-supervision information within the image itself. The reason behind this is that image patches from the same image will automatically have the same class label. This provides an important self-supervision constraint to train the network. However, this one-image approach will limit  the learning capability of the network since the network is not able to 
compare multiple images.
As we know, when human eyes are examining images to determine which features are discriminative, they need to compare multiple images to determine which set of features are consistent across images and which set of features are background noise \cite{gazzaniga2009cognitive}. Therefore, in unsupervised learning, it is highly desirable to utilize the information across images.

\begin{figure}[h]
	\begin{center}
		\includegraphics[width=0.92\linewidth]{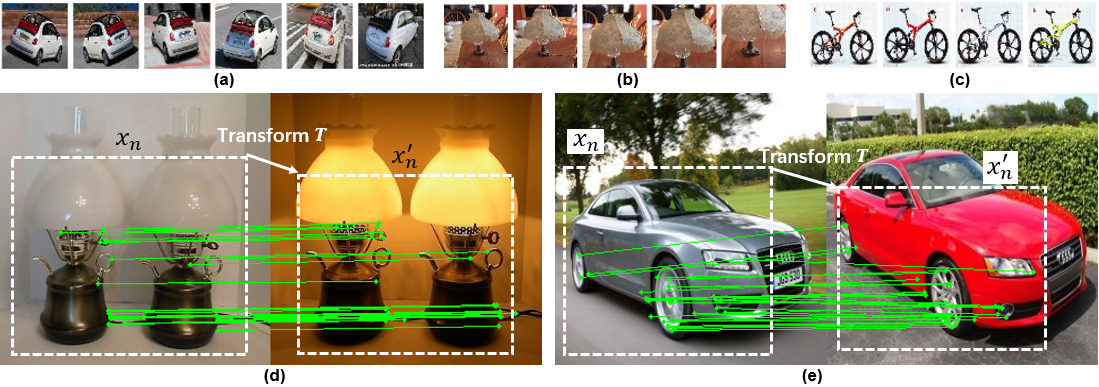}
	\end{center}
	\vspace{-6mm}
	\caption{Sub-image matching for cross-image supervision.}
	\label{fig:matching}
	\vspace{-4mm}
\end{figure}

Figs. \ref{fig:matching}(a)-(c) show image samples from the Cars and SOP benchmark datasets. We can see that images from the same class exhibit strong similarity between images, especially in the object regions. The question is how to utilize these unlabeled images to create reliable self-supervision information for unsupervised learning?
In this work, we propose to perform keypoint or sub-image matching across images. Specifically, as illustrated in Fig. \ref{fig:matching}(d) and (e), for a given image sample $I_n$, in the pre-processing stage, we perform affine-SIFT \cite{morel2009asift} keypoint matching between $I_n$ and other images in the dataset and find the top matches with confidence scores about a very high threshold.  We then crop out the sub-images containing
high-confidence keypoints  as $x_n$ and $x'_n$
which are related by a transform $T$. 
This high-confidence constraint aims to ensure that $x_n$ and $x'_n$ are having the same object class or semantic label. 
In this way, for each image in the $k$-means cluster, we can find multiple high-confidence matched sub-images. This will significantly augment the training set, establish cross-image self-supervision,  and provide significantly enhanced visual diversity for the network to learn more robust and discriminative features. In this work, we combine this cross-image supervision with the transformed attention consistency. Let $\{(u_i, v_i)\}$ and $\{(u'_i, v'_i)\}$, $1\le i\le N$, be the set of matched keypoints in $x_n$ and $x'_n$. 
We wish that, within the small neighborhoods of these matched keypoints, the attention maps $M_n$ and $M'_n$ are consistent. 
To define a small neighborhood around a keypoint $(u_i, v_i)$, we use the following 2-D Gaussian kernel,
\begin{equation}
    \phi(u-u_i, v-v_i) = e^{-\frac{(u-u_i)^2}{2\sigma_u^2} -\frac{(v-v_i)^2}{2\sigma_v^2}}.
\end{equation}
Let 
\begin{equation}
    \Gamma(u, v) = \sum_{i=1}^M \phi(u-u_i, v-v_i), \quad
    \Gamma'(u, v) = \sum_{i=1}^M \phi(u-u'_i, v-v'_i)
\end{equation}
which define two masks to indicate the neighborhood areas around these matched keypoints in these two attention maps. The extended transformed attention consistency becomes
\begin{equation}
    L_{TAC} = \sum_{(u, v)} |M_n(u, v)\cdot \Gamma(u, v)  - M'_n(u, v)\cdot \Gamma'(u, v)|^2,
\end{equation}
which compares the difference between these two attention maps around these matched keypoints.
Compared to the label propagation method developed for semi-supervised learning \cite{zhu2002learning,zhu2005semi}, 
our cross-image supervision method is unique in the following aspects: (1) it discovers sub-images of the same label (with very high probability) from unlabeled images. (2) It establishes the transform between these two sub-images and combines with the transformed attention consistency to achieve efficient unsupervised deep metric learning. 


\begin{figure*}[t]
    \begin{center}
 \hspace{-2mm}\subfigure[CUB]{
			\label{fig:retrieval-cub}
        \includegraphics[width=1\linewidth]{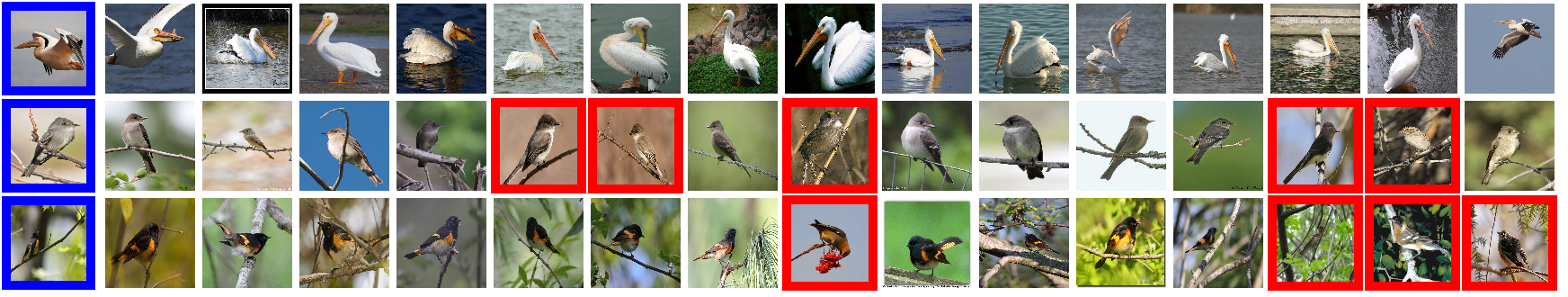}
		}
    \hspace{-1mm}\subfigure[Cars]{
			\label{fig:retrieval-cars}
        \includegraphics[width=1\linewidth]{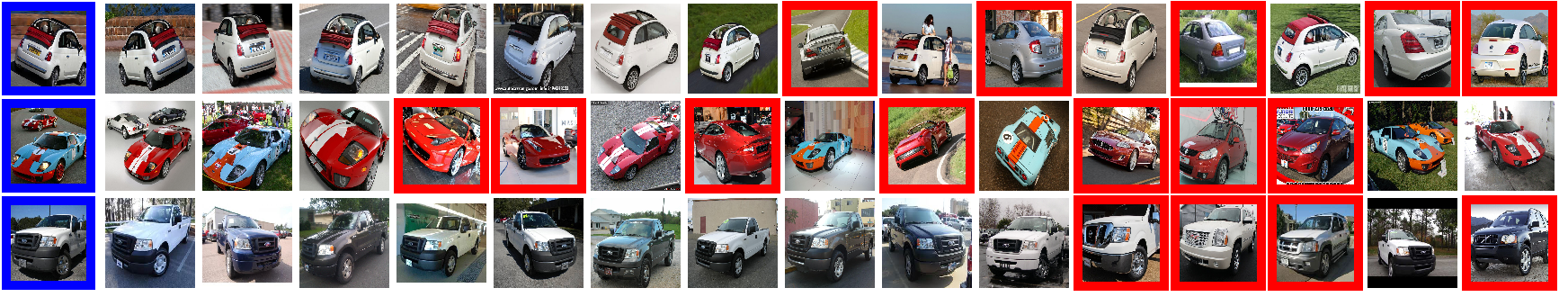}
		}
   \hspace{-0mm}\subfigure[SOP]{
			\label{fig:retrieval-sop}
        \includegraphics[width=1\linewidth]{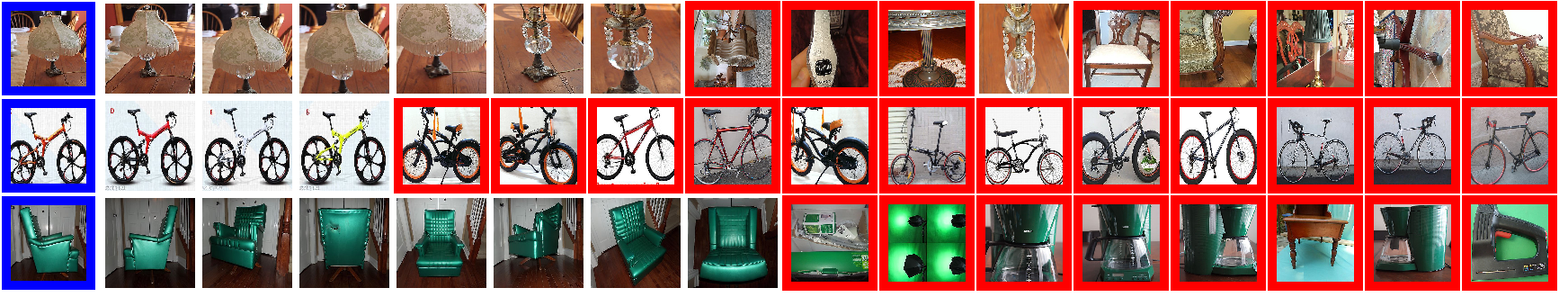}
		}
	    \end{center}
    \vspace{-6mm}
	\caption{Retrieval results of some example queries on CUB, Cars, and SOP datasets. The query images and the negative retrieved images are highlighted with \textcolor{blue}{blue} and \textcolor{red}{red}.}
	\label{fig:example}
	\vspace{-6mm}
\end{figure*}

\section{Experimental Results}
In this section, we conduct extensive experiments on benchmark datasets in image retrieval settings  to evaluate the proposed TAC-CCL method for unsupervised deep metric learning.

\subsection{Datasets}
We follow existing papers on unsupervised deep metric learning \cite{ye2019unsupervised} to evaluate our proposed methods on the following three benchmark datasets.
\textbf{(1) CUB-200-2011 (CUB)} is composed of 11,788 images of birds from 200 classes. The first 100 classes (5864 images) are used for training, with the rest 100 classes (5924 images) for testing.
\textbf{(2) Cars-196 (Cars)} contains 16,185 images of 196 classes of car models. We use the first 98 classes with 8054 images for training, and remaining 98 classes (8131 images) for testing.
\textbf{(3) Stanford Online Products (SOP)} has 22,634 classes (120,053 images) of online products. We use the first 11,318 products (59,551 images) for training and the remaining 11,316 products (60,502 images) for testing. 
The training classes are separated from the test classes.
We use the standard image retrieval performance metric (Recall@k), for performance evaluations and comparisons.

\begin{table}[h]
\footnotesize
		\caption{Recall@\textit{K} (\%) performance on CUB dataset in comparison with other methods}
		\vspace{-2mm}
		\begin{center}
		\resizebox{0.8\linewidth}{!}{
			\begin{tabularx}{9.2cm}{lc|cccc}
				\hline 
				\multirow{2}[1]*{Methods} &\multirow{2}[1]*{Backbone}	&\multicolumn{4}{c}{\textbf{CUB}} \\
				\cline{3-6} 
				& & R@1 & R@2 & R@4 & R@8 \\
				\hline

				\multicolumn{4}{l}{\textbf{\textit{Supervised Methods}}} \\

				\hline

				ABIER \cite{opitz2018deep} & GoogLeNet   & 57.5 & 68.7 & 78.3 & 86.2 \\
				ABE \cite{kim2018attention} & GoogLeNet  & 60.6 & 71.5 & 79.8 & 87.4 \\

                Multi-Similarity \cite{wang2019multi}    &	BN-Inception     & 65.7 & 77.0 & 86.3 & 91.2 \\ 
				\hline
				\multicolumn{4}{l}{\textbf{\textit{Unsupervised Methods}}} \\
				\hline
                Examplar \cite{dosovitskiy2015discriminative} &GoogLeNet & 38.2 & 50.3 & 62.8 & 75.0 \\ 
                NCE \cite{wu2018unsupervised}   &   GoogLeNet           & 39.2 & 51.4 & 63.7 & 75.8 \\
                DeepCluster \cite{caron2018deep}    &   GoogLeNet       & 42.9 & 54.1 & 65.6 & 76.2 \\
                MOM \cite{iscen2018mining}     &    GoogLeNet           & 45.3 & 57.8 & 68.6 & 78.4 \\
                Instance \cite{ye2019unsupervised} & GoogLeNet          & \textcolor{blue}{46.2} & \textcolor{blue}{59.0} & \textcolor{blue}{70.1} & \textcolor{blue}{80.2}  \\
				\hline
				\hline
				\textbf{TAC-CCL (baseline)} & GoogLeNet & \textbf{53.9} & \textbf{66.2} & \textbf{76.9} & \textbf{85.8} \\
               \textbf{TAC-CCL} & GoogLeNet & \textbf{57.5} & \textbf{68.8} & \textbf{78.8} & \textbf{87.2} \\
				\cline{1-6} 
				\multicolumn{2}{c|}{Gain } & \textbf{\textcolor{red}{$+$11.3}} & \textbf{\textcolor{red}{$+$9.8}} & \textbf{\textcolor{red}{$+$8.7}} & \textbf{\textcolor{red}{$+$7.0}} \\
				\hline
			\end{tabularx}
			}
		\end{center}
		\label{tab:cub}
	\end{table}

\subsection{Implementation Details}

We implement our proposed method by PyTorch and follow the standard experimental settings in existing papers \cite{oh2016deep,wang2019multi,ye2019unsupervised} for performance comparison. We use the same GoogLeNet \cite{szegedy2015going} pre-trained on ImageNet as the backbone network  \cite{movshovitz2017no,oh2016deep,sohn2016improved} and 
a CBAM \cite{woo2018cbam} attention module is placed after the \textit{inception\_5b} layer. A fully connected layer is then added on the top of the network as the embedding layer. The default dimension of embedding is set as 512.
For the clustering, we set the number of clusters $K$ to be 100 for the CUB and Cars datasets, and $K=10000$ for the SOP dataset.
For each batch, we follow the data sampling strategy in multi-similarity loss \cite{wang2019multi} to sample 5 images per class.
For data augmentation,  images in the training set are randomly cropped at size 227 $\times$ 227 with random horizontal flipping, while the images in testing set is center cropped. 
Adam optimizer \cite{kingma2014adam} is used in all experiments and the weigh decay is set as $5e^{-4}$.

\begin{table}[h]
\footnotesize
		\caption{Recall@\textit{K} (\%) performance on Cars dataset in comparison with other methods}
        \vspace{-2mm}
		\begin{center}
		\resizebox{0.8\linewidth}{!}{
			\begin{tabularx}{9.0cm}{lc|cccc}
				\hline 
				\multirow{2}[1]*{Methods} &\multirow{2}[1]*{Backbone}	&\multicolumn{4}{c}{\textbf{Cars}} \\
				\cline{3-6} 
				& & R@1 & R@2 & R@4 & R@8  \\ 
				\hline
				\multicolumn{4}{l}{\textbf{\textit{Supervised Methods}}} \\
				\hline
				
                ABIER \cite{opitz2018deep}    &GoogLeNet                 & 82.0 & 89.0 & 93.2 & 96.1 \\
                ABE \cite{kim2018attention}     &  GoogLeNet            & 85.2 & 90.5 & 94.0 & 96.1 \\
				
                Multi-Similarity \cite{wang2019multi}   &BN-Inception       & 84.1 & 90.4 & 94.0 & 96.5 \\
				\hline
				\multicolumn{4}{l}{\textbf{\textit{Unsupervised Methods}}} \\
				\hline

                Examplar \cite{dosovitskiy2015discriminative} &GoogLeNet & 36.5 & 48.1 & 59.2 & 71.0 \\
                NCE \cite{wu2018unsupervised}      &  GoogLeNet         & 37.5 & 48.7 & 59.8 & 71.5 \\              
                DeepCluster \cite{caron2018deep}   &  GoogLeNet         & 32.6 & 43.8 & 57.0 & 69.5 \\
               
                MOM \cite{iscen2018mining}       &   GoogLeNet          & 35.5 & 48.2 & 60.6 & 72.4 \\
                Instance \cite{ye2019unsupervised}    &   GoogLeNet     & \textcolor{blue}{41.3} & \textcolor{blue}{52.3} & \textcolor{blue}{63.6} & \textcolor{blue}{74.9} \\
				\hline
				\hline
				\textbf{TAC-CCL (baseline)} & GoogLeNet & \textbf{43.0} & \textbf{53.8} & \textbf{65.3} & \textbf{76.0} \\
                \textbf{TAC-CCL} & GoogLeNet                       & \textbf{46.1} & \textbf{56.9} & \textbf{67.5} & \textbf{76.7} \\

				\cline{1-6} 
				\multicolumn{2}{c|}{Gain } & \textbf{\textcolor{red}{$+$4.8}} & \textbf{\textcolor{red}{$+$4.6}} & \textbf{\textcolor{red}{$+$3.9}} & \textbf{\textcolor{red}{$+$1.8}} \\
				\hline
			\end{tabularx}
			}
		\end{center}
		\label{tab:cars}
	\end{table}

\begin{table}[h]
\footnotesize
		\caption{Recall@\textit{K} (\%) performance on SOP dataset in comparison with other methods}
        \vspace{-2mm}
		\begin{center}
		\resizebox{0.8\linewidth}{!}{
			\begin{tabularx}{8.8cm}{lc|ccc}
				\hline 
				\multirow{2}[1]*{Methods} 

				&\multirow{2}[1]*{Backbone}
				&\multicolumn{3}{c}{\textbf{SOP}} \\

				\cline{3-5} 
				 & & R@1 & R@10 & R@100  \\
				\hline
				\multicolumn{4}{l}{\textbf{\textit{Supervised Methods}}} \\
				\hline

                ABIER \cite{opitz2018deep}&GoogLeNet                 & 74.2 & 86.9 & 94.0  \\
                ABE \cite{kim2018attention} &  GoogLeNet               & 76.3 & 88.4 & 94.8  \\

                Multi-Similarity \cite{wang2019multi} &BN-Inception      & 78.2 & 90.5 & 96.0  \\
				\hline

				\hline
				\multicolumn{4}{l}{\textbf{\textit{Unsupervised Methods}}} \\
				\hline
				
                Examplar \cite{dosovitskiy2015discriminative} &GoogLeNet   & 45.0 & 60.3 & 75.2 \\
                NCE \cite{wu2018unsupervised}  &  GoogLeNet            & 46.6 & 62.3 & 76.8 \\
                DeepCluster \cite{caron2018deep} &   GoogLeNet        & 34.6 & 52.6 & 66.8 \\
                MOM \cite{iscen2018mining} &   GoogLeNet                & 43.3 & 57.2 & 73.2 \\
                Instance \cite{ye2019unsupervised} &    GoogLeNet       & \textcolor{blue}{48.9} & \textcolor{blue}{64.0} & \textcolor{blue}{78.0} \\
				\hline
				\hline

                \textbf{TAC-CCL (baseline)} & GoogLeNet & \textbf{62.5} & \textbf{76.5} & \textbf{87.2} \\
                \textbf{TAC-CCL} & GoogLeNet & \textbf{63.9} & \textbf{77.6} & \textbf{87.8} \\
				\cline{1-5} 
				\multicolumn{2}{c|}{Gain}  &  \textbf{\textcolor{red}{$+$15.0}} & \textbf{\textcolor{red}{$+$13.6}} & \textbf{\textcolor{red}{$+$9.8}} \\
				\hline
			\end{tabularx}
			}
		\end{center}
		\label{tab:sop}
	\end{table}

\subsection{Performance Comparisons with State-of-the-Art Methods}

We compare the performance of our proposed methods with the state-of-the-art unsupervised methods on image retrieval tasks. The mining on manifolds (MOM) \cite{iscen2018mining} and the invariant and spreading instance feature method (denoted by Instance) \cite{ye2019unsupervised} are current state-of-the-art methods for unsupervised metric learning.
They both use the  GoogLeNet \cite{szegedy2015going} as the backbone encoder. 
In the Instance paper \cite{ye2019unsupervised}, 
the authors have also implemented three other state-of-the-art methods originally developed for supervised deep metric learning and adapted them to unsupervised metric learning tasks:
Examplar \cite{dosovitskiy2015discriminative}, NCE (Noise-Contrastive Estimation) \cite{wu2018unsupervised}, and DeepCluster \cite{caron2018deep}. We include the results of these methods for comparisons. 
We have also included the performance of recent supervised deep metric learning methods for comparison so that we can see the performance difference between unsupervised metric learning and supervised one. 
These methods include: ABIER \cite{opitz2018deep}, and  ABE \cite{kim2018attention}, and MS (Multi-Similarity) \cite{wang2019multi}. Both ABIER and ABE methods are using the GoogleNet as the backbone encoder. The MS method is using the BN-Inception network \cite{ioffe2015batch} as the backbone encoder.

The results for the CUB, Cars, and SOP datasets are summarized in Tables \ref{tab:cub}, \ref{tab:cars}, and \ref{tab:sop}, respectively. 
We can see that our proposed TAC-CCL method achieves new state-of-the-art performance in unsupervised metric learning on both fine-grained CUB and Cars datasets and the  large-scale SOP dataset. On the CUB dataset, our TAC-CCL improves the Recall@1 by 11.3\% and is even competitive to some supervised metric learning methods, e.g.,  ABIER \cite{opitz2018deep}. On the Cars dataset, our TAC-CCL outperforms the current state-of-the-art Instance method \cite{ye2019unsupervised} by 4.8\%. On SOP, our  method achieves 63.9\% and outperforms existing methods by a large margin of 15\%. 
For other Recall@K rates with large values of $k$, the amount of improvement is also very significant. 
Note that our baseline system achieves a large improvement over existing methods. The proposed TAC-CCL approach further improves upon this baseline system by another 1.4-3.6\%.

Fig. \ref{fig:example} shows examples of retrieval results from the  CUB, Cars, and SOP datasets. 
In each row, the first image highlighted with a blue box is the query image. The rest images are the top 15 retrieval 
results. Images highlighted with red boxes are from different classes. It should be noted that some classes have very small number of samples.  
We can see that our TAC-CCL can learn discriminitive features to achieve satisfying retrieval results, even for these challenging tasks. For example, at the first row of the SOP dataset, our model is able to learn the glass decoration feature under the lampshade, which is a unique feature of the query images. In addition, the negative retrieved results are also visually closer to the query images.

\subsection{Ablation Studies}

In this section, we conduct ablation studies  to perform  in-depth analysis of our proposed method and its different components.

\begin{figure}[h]
	\begin{center}
		\includegraphics[width=\linewidth]{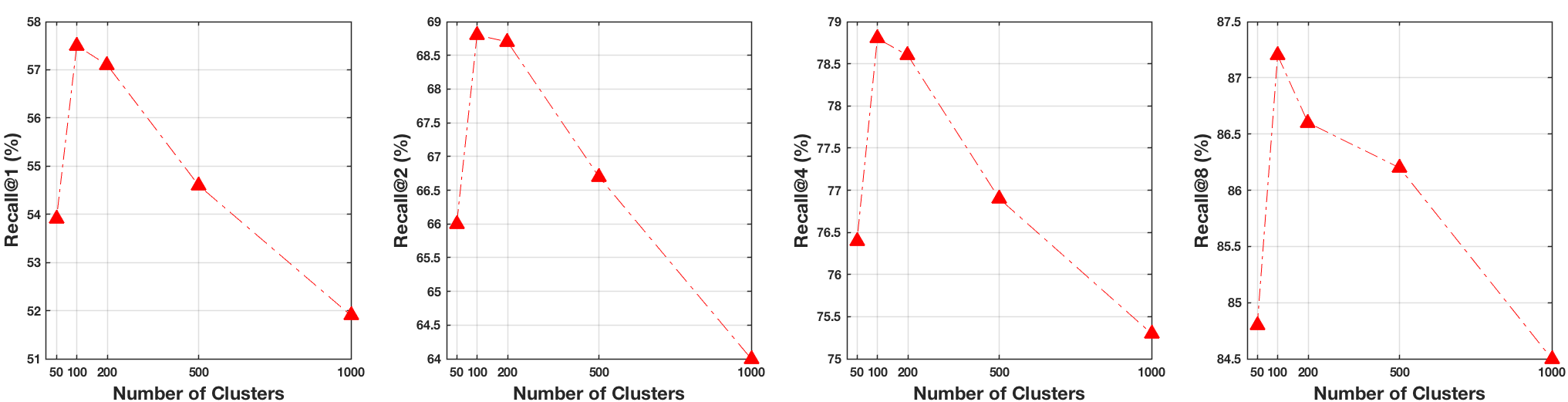}
	\end{center}
	\vspace{-6mm}
	\caption{Recall@\textit{K} (\%) performance on CUB dataset in comparison with different number of clusters}
	\label{fig:cluster}
	\vspace{-4mm}
\end{figure}

\begin{figure}[h]
	\begin{center}
		\includegraphics[width=\linewidth]{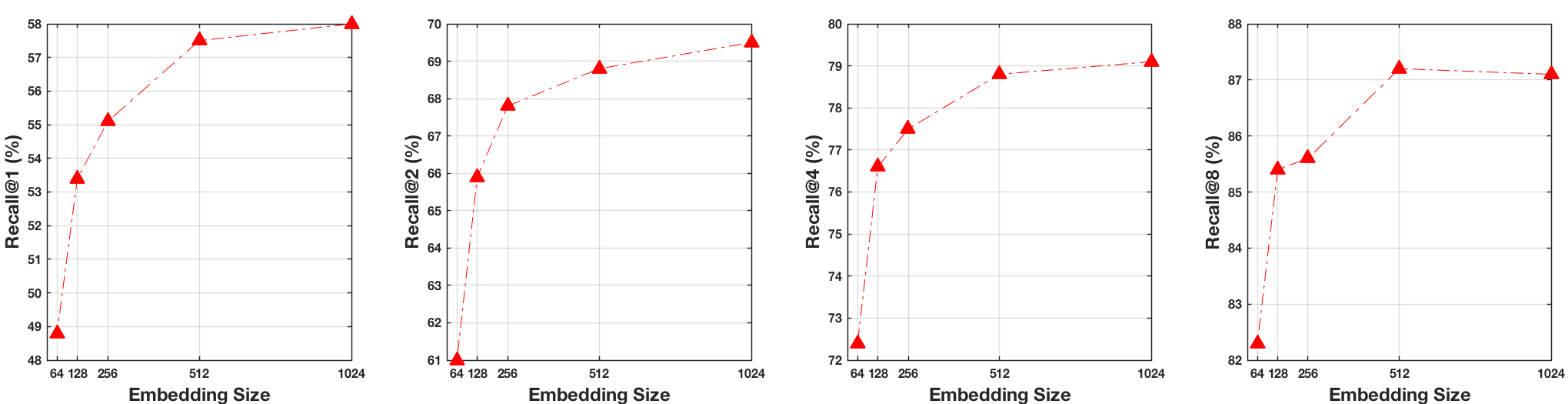}
	\end{center}
	\vspace{-6mm}
	\caption{Recall@\textit{K} (\%) performance on CUB dataset in comparison with different embedding sizes}
	\label{fig:embedding}
	\vspace{-4mm}
\end{figure}

\textbf{(1) Impact of the number of clusters.}
The proposed contrastive clustering loss is based on clustering in the feature space. 
The number of clusters $K$ is a critical parameter for the proposed method since it determines the number of pseudo labels. We conduct the following ablation study experiment on the CUB data to study the impact of $K$. 
The first plot in Fig. \ref{fig:cluster} shows the Recall@1 results with different values of $K$: 50, 100, 200, 500, and 1000. The other three plots show the results for Recall@2, 4, and 8. We can see that, on this dataset, the best value of $K$ is 100, which is the number of test classes in the CUB dataset.
The performance drops when $K$ increases. 
This study suggests that the best value of $K$ is close to the truth number test classes of the dataset.

\textbf{(2) Impact of different embedding sizes.} 
In this ablation study,  we follow existing supervised metric learning methods \cite{wang2019multi,opitz2018deep} to study the impact of different embedding sizes, or the size of the embedded feature. For example, the feature size ranges from 64, 128, 256, 512, to 1024.
The first plot of Fig.  \ref{fig:embedding} shows the Recall@1 results for different embedding size. The results for Recall@2, 4, 8 are shown in the other three plots.
We can see that unsupervised metric learning performance increases with the embedding size since it contains more feature information with enhanced discriminative power.

\textbf{(3) Performance  contributions  of  different  algorithm  components.}  Our proposed system has three major components: the baseline system for unsupervised deep metric learning, transformed attention consistency (TAC), and contrastive clustering loss (CCL).
In this ablation study, we aim to identify the contribution of each algorithm component on different datasets.  Table \ref{ablation:all} summarizes the performance results on the CUB, Cars, and SOP datasets using three different method configurations: (1) the baseline system, (2) baseline with CCL, and (3) baseline with CCL + TAC. 
We can see that both the CCL and TAC approaches significantly improve the performance. 

In our Supplemental materials, we will provide additional method implementation details, experimental results,  and ablation studies. 

\begin{table*}[h]
		\caption{The performance of different components from our TAC-CCL method on CUB, Cars, and SOP datasets.}
        \vspace{-2mm}
		\begin{center}
		\resizebox{0.95\linewidth}{!}{
			\begin{tabularx}{10.5cm}{c|cccc||cccc||ccc}
				\hline 
				 \multirow{2}[1]*{}	&\multicolumn{4}{c||}{\textbf{CUB}} &\multicolumn{4}{c||}{\textbf{Cars}}&\multicolumn{3}{c}{\textbf{SOP}}\\
				\cline{2-12} 
				& R@1 & R@2 & R@4 & R@8  & R@1 & R@2 & R@4 & R@8  & R@1 & R@10 & R@100 \\
				\hline
				\hline
				Baseline  & 53.9 & 66.2 & 76.9 & 85.8  & 43.0 & 53.8 & 65.3 & 76.0    & 62.5 & 76.5 & 87.2 \\
				$+$CCL & 55.7 & 67.8 & 77.5 & 86.2  & 44.7 & 55.6 & 65.9 & 75.7 & 63.0 & 76.8 & 87.2 \\
				$+$TAC & \textbf{57.5} & \textbf{68.8} & \textbf{78.8} & \textbf{87.2}  & \textbf{46.1} & \textbf{56.9} & \textbf{67.5} & \textbf{76.7}  & \textbf{63.9} & \textbf{77.6} & \textbf{87.8} \\

				\hline
			\end{tabularx}
			}
		\end{center}
		\label{ablation:all}
		\vspace{-8mm}
	\end{table*}
\section{Conclusions}

In this work, we have developed a new approach to unsupervised deep metric learning based on image comparisons, transformed attention consistency, and constrastive clustering loss. This transformed attention consistency leads to a pairwise self-supervision loss, allowing us to learn a Siamese deep neural network to encode and compare images against their transformed or matched pairs. To further enhance the inter-class discriminative power of the feature generated by this network, we have adapted the concept of triplet loss from supervised metric learning to our unsupervised case and introduce the contrastive clustering loss. Our extensive experimental results on  benchmark datasets demonstrate that our proposed method outperforms current state-of-the-art methods by a large margin.

\clearpage
%
%
\bibliographystyle{splncs04}
\bibliography{egbib}
\end{document}